\documentclass[11pt,a4paper]{article}
\usepackage[T1]{fontenc}
\usepackage[utf8]{inputenc}
\usepackage[english]{babel}
\usepackage[margin=1in]{geometry}
\usepackage{lmodern}
\usepackage{microtype}
\usepackage{amsmath,amsfonts,amssymb}
\usepackage{graphicx}
\usepackage{booktabs,array,multirow}
\usepackage{csquotes}
\usepackage[backend=biber,style=apa]{biblatex}
\usepackage{hyperref}
\hypersetup{hidelinks,pdftitle={Benchmarking LLM Next-Step Recommendations with Knowledge Space Theory},pdfauthor={Peter Steiner; Jan Hochweber}}
\title{Benchmarking LLM Next-Step Recommendations with Knowledge Space Theory}
\author{%
Peter Steiner\thanks{ORCID: \href{https://orcid.org/0009-0001-5491-9464}{0009-0001-5491-9464}} \quad
Jan Hochweber\thanks{ORCID: \href{https://orcid.org/0000-0002-7855-2097}{0000-0002-7855-2097}}\\[0.6em]
{\small Institute of Educational Psychology}\\
{\small St.Gallen University of Teacher Education (PHSG)}\\
{\small St.Gallen, Switzerland}}
\date{}
\begin{document}
\maketitle
\begin{abstract}
Selecting an LLM for next-step recommendation requires benchmarks that reflect the intended educational task and domain. We present a benchmarking method that uses Knowledge Space Theory (KST) as a formal evaluation reference. Given a task catalogue and an existing prerequisite relation, the method derives admissible knowledge states and the complete set of immediate next steps for each state. Recommendations generated by a model under a specified prompt are compared with these reference sets. A sample evaluation study uses 2,400 responses from six models across four mathematics domains, four prompt conditions, and two catalogue orders. Exact-set accuracy was 61.3\% with descriptions only, 62.8\% when models were asked to reconstruct prerequisites, 96.3\% with supplied prerequisite relations, and 98.2\% with the complete knowledge space. Performance varied across models and domains, and recommendation sets agreed across the two presentations in 85.25\% of pairs. The benchmark supports comparisons of model--prompt combinations against a context-specific reference. Its scores describe structural recommendation accuracy under the evaluated conditions; they do not establish learning benefit or the necessity of using an LLM for selection.
\end{abstract}

\subsection*{Notes for Practice}
\begin{itemize}
\item The structural accuracy of LLM-based learning recommendations can be validated with Knowledge Space Theory.
\item No single best model emerged; rankings changed with the prompt condition and the content domain.
\item When prerequisite relations and the learner’s knowledge state are available, admissible next steps can be computed deterministically. An LLM may explain these options, but explanation quality and learning benefit require separate evaluation.
\end{itemize}

\section{Introduction}

Building LLM-based digital learning tools presents a practical selection problem: Which large language model (LLM) and with which prompt is suitable for a particular educational task? For learning recommendation, a relevant capability is selecting content that fits a learner's current knowledge and the prerequisite relations of underlying content domain. A general model score offers limited guidance about this requirement. A task-specific benchmark can instead estimate how often a candidate model with a specific prompt recovers the complete set of structurally admissible next learning steps. Such evidence connects the decision-support aims of learning analytics \parencite{siemens2012learninganalytics} with evidence for model and prompt selection.

LLMs appear especially attractive for recommendation tasks. They can interpret competency descriptions, produce accessible explanations, and be integrated easily without training a domain-specific generator. Recent work has explored LLMs as learning assistants and generators of research-informed formative feedback \parencite{steinert2024learningassistants,kasneci2023chatgpt}. Yet, a fluent recommendation can conceal a structural error. A model may treat two tasks as parallel when a given curriculum defines one as prerequisite to the other, omit feasible next tasks, or recommend an already mastered task. Evaluations based on perceived plausibility or text quality do not directly establish whether a generated recommendation fits the intended progression.

The validity of a learning recommendation depends on its context. Some prerequisites reflect the mathematical operations required by a task: for example, determining a linear function's equation from its graph involves identifying its slope and intercept. Other dependencies reflect the instructional sequence adopted by a curriculum or classroom. A benchmark for recommendation therefore needs an explicit account of the progression intended in a specific setting. This allows the same evaluation procedure to be applied to different content domains in different educational context without treating one instructional sequence as universally correct.

Knowledge Space Theory (KST) provides an evaluation reference for this problem. It accounts for a prerequisite structure within a domain. Hence, it can be used to provide a theoretical framework to define learning states and next feasible learning steps from a given learning state \parencite{doignon1985spaces,falmagne2011learning}. An LLM-based learning recommendation can then be compared with a KST-derived one.

The contribution of this paper is a KST-based benchmarking method for assessing the structural accuracy of LLM learning recommendations under specified domain and prompt conditions. An empirical application illustrates the method using four existing mathematics domains, six models, four prompt conditions that vary the provision or elicitation of structural information, and two component orders. This application demonstrates how the benchmark can inform practical model and prompt selection for a particular domain.

\section{Theoretical Background}

\subsection{Benchmarking LLMs for Application-Specific Model Selection}
\label{sec:related-benchmarking}

Benchmarks provide common tasks and scoring procedures for comparing LLMs, but their usefulness for model selection depends on the relationship between the evaluated tasks and the intended application. MMLU assesses knowledge and problem-solving performance across 57 academic and professional subjects \parencite{hendrycks2021measuring}. Such broad coverage supports comparisons of general capabilities, while application-specific evaluation requires a more explicit account of the relevant tasks and criteria. Holistic Evaluation of Language Models (HELM) addresses this need by organizing evaluation around scenarios, adaptation procedures, and multiple metrics, including accuracy, robustness, and efficiency \parencite{liang2023holistic}. This perspective makes model suitability a question of performance under specified conditions rather than a property captured by a single aggregate score.

The interpretation of benchmark scores also raises questions of validity. \textcite{raji2021everything} criticize the assumption that performance on influential benchmarks can stand in for broadly generalizable capabilities. More recently, \textcite{gevers2025benchmarks} distinguish threats to internal, external, statistical, and construct validity in LLM benchmarking and propose complementing standardized benchmarks with targeted test suites for downstream use cases. For educational model selection, this distinction is consequential: answering subject-matter questions and selecting appropriate next learning steps are different tasks. Evidence for the former does not directly establish competence in the latter.

Behavioral testing offers a complementary approach to evaluation. CheckList organizes tests around capabilities and expected input--output behaviors, including invariance under transformations that should not change the answer \parencite{ribeiro2020checklist}. Its contribution is to make specific failures visible that an aggregate accuracy score can conceal. Applied to next-step recommendation, this perspective motivates testing explicitly defined requirements: respecting prerequisites, excluding mastered components, returning all admissible alternatives, and preserving the recommendation set when the same information is presented in a different order.

Prompting and scoring are themselves part of the evaluation procedure. \textcite{sclar2024formatting} demonstrate substantial sensitivity to meaning-preserving formatting changes in few-shot evaluations and show that formats need not affect different models similarly. Consequently, a comparison under one arbitrarily selected format may provide an incomplete account of relative performance. However, \textcite{hua2025flaw} show that some apparent prompt sensitivity arises from evaluation artifacts, including scoring procedures that reject semantically equivalent answers. Together, these findings motivate documenting prompt conditions, examining presentation robustness, and separating changes in the selected answer from differences in its verbal expression.

Educational benchmarks increasingly operationalize capabilities beyond subject knowledge. MathTutorBench distinguishes mathematical expertise, understanding of student solutions, and pedagogical response generation, using a learned reward model to assess open-ended scaffolding \parencite{macina2025mathtutorbench}. Its results indicate that stronger problem-solving performance does not automatically imply stronger tutoring performance. The BEA 2025 shared task likewise evaluates systems that assess tutor responses against human annotations for mistake identification, mistake location, guidance, and feedback actionability \parencite{kochmar2025findings}. These contributions establish educationally specific evaluation targets, while focusing primarily on dialogue responses and their pedagogical properties.

The present work complements these approaches with a formal reference for one defined recommendation task. Given an existing domain structure and a knowledge state, Knowledge Space Theory determines the complete set of admissible immediate next steps. A chosen model--prompt combination can be evaluated against this set regardless of whether the prompt also supplies the prerequisite relations. Mapping outputs to component identifiers permits deterministic set comparison without judging the wording of their explanations. The resulting estimates describe structural recommendation accuracy conditional on the domain, state coverage, and prompt. This provides EdTech providers and researchers with evidence for application-specific model selection, while leaving pedagogical dialogue quality and learning outcomes as distinct evaluation targets.

\subsection{From Learner State to Actionable Feedback}

Feedback is formative when it helps close a gap between current and desired performance and supports subsequent action \parencite{hattie2007feedback,shute2008formative}. Meta-analytic evidence shows that effects depend on the information and level of feedback rather than on feedback presence alone \parencite{wisniewski2020feedback}. In adaptive learning, the analogous design problem is selecting an activity that is attainable from the learner's current knowledge state. A recommendation that ignores prerequisites may ask learners to attempt tasks for which they are not yet prepared. Omitting admissible alternatives may also unnecessarily narrow their available choices. The benchmark operationalizes this structural requirement; it does not evaluate the formative effect of the resulting feedback.

Intelligent tutoring systems address this problem by coordinating an explicit domain model, a learner model, and pedagogical rules. The Knowledge-Learning-Instruction framework similarly emphasizes mappings among knowledge components, learning events, and instruction \parencite{koedinger2012kli}. By contrast, a general-purpose LLM contains no guaranteed representation of the local curriculum structure. Prompting can add context, but the content and form of that context become part of the learning-analytics design.

This distinction is visible in recent LLM studies. Research-based prompt architectures can elicit multiple forms of cognitive and metacognitive scaffolding \parencite{steinert2024learningassistants}. However, systematic prompt design does not by itself verify whether a suggested activity is feasible for a particular learner state. Borchers and Shou \parencite{borchers2025adaptivity} varied tutoring-system context in 75 authentic scenarios and found that current LLMs only marginally reproduced ITS-like adaptivity. Their result motivates component-wise prompt evaluations: rather than treating a long context as one intervention, researchers should test which learner- and domain-model elements change the generated action.

LLM use also raises reliability and accountability concerns. Educational applications inherit risks of hallucination and unsupported inference \parencite{ji2023hallucination}, while reviews identify privacy, bias, overreliance, transparency, and uneven pedagogical validity as recurring challenges \parencite{yan2024challenges}. The present benchmark addresses a specific component of this broader evaluation problem: agreement between a generated recommendation set and the next steps defined by an existing domain model. It provides evidence for system selection without assuming that structural agreement captures every requirement of educational use.

\section{The KST-Based Benchmarking Method}
\label{sec:benchmarking-method}

The proposed method evaluates whether recommendations generated by a model under a specified prompt match the admissible immediate next steps in a domain with a specific underlying prerequisite relation. Its reference inputs are a catalogue of tasks and a prerequisite relation among these tasks that expresses the intended progression. From these inputs, the benchmark derives admissible learning states and their reference recommendation sets, selects the states to evaluate, obtains model recommendations, and scores their agreement with the reference. The method can therefore be reused with different domain structures while retaining the same task definition and scoring procedure.

\subsection{KST and Outer Fringes}

Let $Q$ be a finite domain of tasks. A surmise relation $\preceq$ on $Q$ is reflexive and transitive. We write $p\preceq q$ when mastery of $q$ implies mastery of $p$. Thus, for distinct tasks, $p$ is a prerequisite of $q$. The relation induces the knowledge space
\begin{equation}
\mathcal{K}_{\preceq}=\{K\subseteq Q : \forall p,q\in Q,\ (q\in K\ \text{and}\ p\preceq q)\ \Rightarrow\ p\in K\}.
\label{eq:states}
\end{equation}
Its states are precisely the subsets closed under prerequisites, including $\emptyset$ and $Q$ \parencite{doignon1985spaces,falmagne2011learning}. In simple terms, it represents all possible combinations of tasks with regard to the prerequisites. The individual sets within the knowledge space are called knowledge state. Hence, the knowledge states are possible learning states in a domain and the knowledge space is the collection of all possible learning states in a domain.

For any knowledge space $\mathcal{K}$ and state $K\in\mathcal{K}$, the reference recommendation set is defined as the outer fringe:
\begin{equation}
K^{\mathcal{O}}=\{q\in Q\setminus K : K\cup\{q\}\in\mathcal{K}\}.
\label{eq:fringe}
\end{equation}
It contains all unmastered tasks that can be added individually without violating the prerequisite structure. In the acyclic prerequisite structures used here, this is equivalent to requiring that all prerequisites of a candidate task have already been mastered. The full state $Q$ has an empty outer fringe. 

% Prerequsite structures can originate from expert elicitation, data, or combinations of both; the QUERY procedure is an established elicitation approach \parencite{kambouri1994query}, and competence-based approaches distinguish problems from underlying skills \parencite{duntsch1995skills}. These are possible sources of benchmark inputs. Their provenance and prerequisite interpretation need to be documented because the benchmark measures agreement with the supplied structure. Creating or validating that structure is a separate activity from evaluating model recommendations.

\subsection{Benchmark Cases and Model Queries}

State coverage is part of the benchmark specification. A domain with $n$ mutually unrelated tasks has $2^n$ admissible learning states, so exhaustive evaluation can become computationally and financially impractical. Larger domains may require a documented sample of admissible states. The selected states and their weights determine whether an aggregate score describes all admissible states, a particular subset, or a target distribution of learning states. For a known prerequisite relation, each selected state's fringe can be computed by checking prerequisites without enumerating the entire knowledge space.

For each selected learning state, the candidate model receives the task catalogue and the mastered-task set through the prompt being assessed. The task is to return all and only admissible immediate next steps. Responses are mapped to canonical task identifiers and compared as unordered sets with the precomputed fringe. The evaluator always has access to the reference structure; the model may receive only task descriptions and the current state, or additionally receive prerequisite relations or admissible states. Holding the reference, state coverage, task, and scoring procedure fixed permits comparisons across models and prompt variants. Output validation and the treatment of invalid responses and technical failures must also be specified so that formatting or collection failures do not silently change the evaluated population.

\subsection{Evaluation Metrics}
\label{sec:metrics}

To evaluate the ability of an LLM to recommend next feasible learning steps, its recommendations are compared with the outer fringe derived from the reference knowledge structure. For a given knowledge state $K$, let $P$ denote the set of problems recommended by the model and $T=K^{\mathcal{O}}$ the corresponding reference set. Differences between these sets can arise in two ways: the model may recommend problems that do not belong to the outer fringe, or it may omit problems that constitute feasible next learning steps.

A direct measure of disagreement is the cardinality of the symmetric difference \parencite{gower1986dissimilarity}:
\begin{equation}
	D(P,T)=|(P\setminus T)\cup(T\setminus P)|.
\end{equation}
This distance counts the additional and omitted recommendations. It equals zero when the two sets coincide and increases by one for each problem on which they differ.

To express agreement relative to the size of the compared sets, Jaccard similarity divides the number of shared recommendations by the number of problems included in either set:
\begin{equation}
	J(P,T)=\frac{|P\cap T|}{|P\cup T|}.
\end{equation}
A value of one indicates complete agreement, whereas zero indicates that the two nonempty sets have no elements in common. The corresponding Jaccard distance is $1-J(P,T)$. The sample study reports similarity, so higher values consistently indicate better performance.

The comparison can also be understood as a precision--recall classification problem \parencite{vanrijsbergen1979information,sokolova2009measures}. Problems recommended by the model that belong to the reference fringe are true positives. Additional recommendations are false positives, and omitted feasible problems are false negatives. Precision describes the proportion of the model's recommendations that are admissible, whereas recall describes the proportion of admissible next steps recovered by the model. Their harmonic mean is the F1 score:
\begin{equation}
	F_1(P,T)=\frac{2|P\cap T|}{|P|+|T|}.
\end{equation}
These measures distinguish between recommending inadmissible problems and failing to identify all available alternatives.

Finally, the perfect-match rate, referred to here as exact-set accuracy, describes the proportion of responses in which the model's recommendation set coincides exactly with the reference outer fringe. It is the strictest criterion: a response is counted as correct only if it contains all feasible next steps and no additional recommendations. Unlike the measures of partial agreement, it does not distinguish between responses with one error and responses with several errors.

For terminal knowledge states, the reference fringe is empty. An empty model recommendation is therefore fully correct. When both sets are empty, exact correctness, precision, recall, F1, and Jaccard similarity are assigned a value of one. An empty prediction for a nonempty reference receives zero on all measures. For a nonempty prediction against an empty reference, recall is one because no reference element is omitted; the remaining measures are zero. The sample study averages the response-level scores, so macro F1 is the mean of the individual F1 scores.

The same set comparisons can be used to examine agreement between recommendations generated under different presentations of the same input. In that case, the two predictions are compared with each other rather than with the reference fringe. Such agreement describes consistency, not correctness: two predictions can coincide while both deviate from the reference.

A code template for applying the method can be found in the supplementary materials: \href{https://github.com/pstnr/Benchmarking-LLM-Next-Step-Recommendations-with-Knowledge-Space-Theory}{GitHub repository}.

\section{A Sample Evaluation Study}
\label{sec:study}

The sample study applies the benchmark to the selection of an LLM and prompt for learning recommendations in linear functions within a Swiss secondary-education context. Its purpose is to illustrate how the method answers application-specific evaluation questions using existing domain structures. Two research questions guide the study:

\begin{description}
\item[RQ1:] How does the structural information provided in the prompt influence recommendation accuracy?
\item[RQ2:] How does recommendation performance vary across models and domains, and how consistent are recommendations across two presentations of the same component catalogue?
\end{description}

\subsection{Study Domains}

The application used four existing linear-function domains with prerequisite relations from the SaTiM project (Table~\ref{tab:domains}). The four structures contain 23 tasks and 50 domain-specific learning states, including one terminal state per domain in which all tasks are mastered. All states were included and weighted equally within the aggregate condition comparisons. 

\begin{table}[hbt]
	\caption{Study domains and exhaustive state coverage}
	\centering
	\small
	\begin{tabular}{lrrr}
		\toprule
		Domain & Components & States & Terminal states \\
		\midrule
		Linear Modelling & 7 & 23 & 1 \\
		Slope & 7 & 11 & 1 \\
		Graph to Equation & 5 & 9 & 1 \\
		Equation to Graph & 4 & 7 & 1 \\
		\midrule
		Total & 23 & 50 & 4 \\
		\bottomrule
	\end{tabular}
	\label{tab:domains}
\end{table}

\subsection{Prompt Conditions and Response Format}

The prompt core was shared across conditions; the structure block and P1's additional graph-output requirement differed (Table~\ref{tab:conditions}). P0 contained no structural block. P1 asked the model to infer and return direct prerequisite relations before applying them. P2 supplied the reference direct cover relations and stated that prerequisite chains are transitive. P3 supplied all and only admissible states and defined the one-component extension rule in Equation~\ref{eq:fringe}. The true outer fringe was never included in any prompt.

\begin{table}[hbt]
	\caption{Experimental structural-information conditions}
	\centering
	\small
	\begin{tabular}{p{0.10\linewidth}p{0.25\linewidth}p{0.55\linewidth}}
		\toprule
		Code & Information source & Condition-specific input or task \\
		\midrule
		P0 & Descriptions only & Infer next steps from component semantics and the mastered set; no structure is supplied. \\
		P1 & Model reconstruction & Infer direct prerequisites from descriptions, return that graph, and use it for recommendations. \\
		P2 & Reference graph & Use supplied direct cover relations and their transitive implications. \\
		P3 & Admissible states & Use the complete list of admissible states and the one-item extension rule. \\
		\bottomrule
	\end{tabular}
	\label{tab:conditions}
\end{table}

Responses were requested as JSON objects under a provider-level schema and checked again locally. The common schema contained a recommendation array of temporary position and justification pairs plus a mutually constrained no-recommendation reason. P1 additionally returned an array of directed position pairs. Duplicate or out-of-range positions, malformed JSON, missing explanations, refusals, extra prose, and truncation were predefined as invalid. A returned invalid response would count as incorrect in the primary scoring; a transport failure without model output would be technical missingness rather than an empty prediction. Complete task descriptions, prerequisite relations, and example prompts for all four conditions are provided in the supplementary repository.

\subsection{Models, Counterbalancing, and Collection}

The dataset contains responses from six API-accessible models collected in September 2026: GPT-5.6 Sol, GPT-5.6 Terra, Claude Fable 5, Claude Sonnet 5, Gemini 3.1 Pro Preview, and Gemini 3.7 Flash. The panel included a higher-capability and a price-performance model from each of three providers. Provider-standard sampling settings were retained, and the output allowance was 16,000 tokens to prevent reasoning-related truncation observed during pre-study testing. Model comparisons describe these recorded versions and settings.

Temporary 1-based positions replaced canonical component IDs in every prompt. For each domain, two deterministic random permutations were generated from SHA-256 seeds and reused across states, conditions, and models. Mastered sets, P2 relations, P3 states, and responses were mapped through the same permutation. These calls compare recommendations across two semantically equivalent presentations. They are not identical-prompt repeats, so disagreement may reflect both presentation effects and stochastic variation.

The fully crossed study design comprised

\begin{equation}
50\text{ states}\times4\text{ conditions}\times6\text{ models}\times2\text{ orders}=2{,}400\text{ calls}.
\end{equation}

Within each model, the 400 study calls were deterministically shuffled so that domain, state, condition, and order were interspersed over collection time. Requests were collected on September 8--9, 2026. A daily quota initially delayed 160 Gemini 3.1 Pro Preview cells; they were resumed after reset with unchanged run identifiers and prompts. Local transport events and every API attempt were retained separately for audit. The final study dataset contains 2,400 API responses, all of which passed schema validation; no response was truncated and no technical-missing placeholder remained. Total estimated API cost, including three provider smoke tests, was US\$34.98.

\subsection{Analysis and Aggregation}

Each of the 50 unique knowledge states received equal weight in the condition comparisons. Within a state and condition, outcomes were averaged across six models and two orders; condition contrasts were then calculated as paired differences across states. Percentile 95\% intervals used 10,000 bootstrap samples of the 50 states. Two-sided randomization tests used 100,000 independent sign flips of the state-level differences. The three adjacent exact-set contrasts were adjusted by Holm's procedure \parencite{holm1979simple}. The condition contrasts were specified by the design, but no single inferential outcome hierarchy had been frozen before analysis; p values are therefore interpreted as controlled descriptive evidence rather than as a preregistered confirmatory family.

Model, domain, precision, recall, F1, and Jaccard results are descriptive. Model comparisons are not treated as causal effects of capability tier because provider, architecture, release status, and pricing differ jointly. Within each domain, every included state receives equal weight. In the pooled condition estimates, domains contribute in proportion to their state counts; these estimates do not represent equal-domain weighting or a population distribution of learners. RQ2 additionally compares the two predictions for each model--condition--state combination after mapping both to canonical identifiers.

The reported evaluation uses the completed 2,400-response dataset. It includes neither additional domains nor identical-prompt repeat measurements or a qualitative evaluation of explanations. The analysis code, extracted recommendation sets, domain specifications, and reference knowledge states are available in the supplementary repository. 

\subsection{Results}

\subsubsection{RQ1: Recommendation Accuracy Across Prompt Conditions}

\begin{figure}[ht]
\centering
\includegraphics[width=0.95\linewidth]{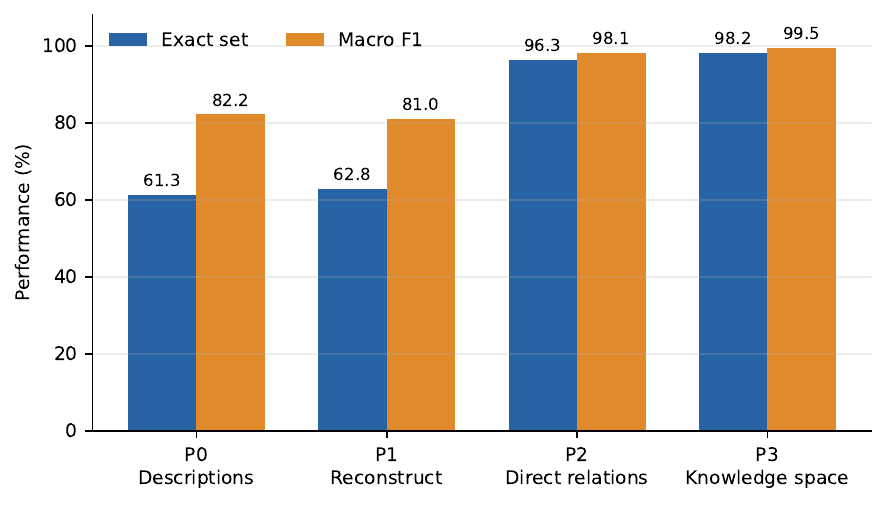}
\caption{Exact-set accuracy and macro F1 across structural-information conditions. Each condition contains 600 responses: 50 states, six models, and two component orders.}
\label{fig:condition}
\end{figure}

All 2,400 responses were schema-valid and scorable, with no technical missingness in the completed dataset. Across all responses, exact-set accuracy was 79.7\%, macro precision 89.6\%, recall 92.9\%, F1 90.2\%, and Jaccard similarity 87.6\%. All 192 terminal-state responses correctly returned an empty recommendation set. Accuracy differed across structural-information conditions (Figure~\ref{fig:condition}): exact accuracy was 61.3\% in P0 and 62.8\% in P1, compared with 96.3\% in P2 and 98.2\% in P3. Macro F1 increased from 81.0\% in P1 to 98.1\% in P2 (Table~\ref{tab:conditionresults}).

\begin{table}[hbt]
\caption{Recommendation performance by structural-information condition}
\centering
\small
\begin{tabular}{lrrrrrr}
\toprule
Condition & $n$ & Exact & Precision & Recall & F1 & Jaccard \\
\midrule
P0 & 600 & 61.3\% & 80.4\% & 88.0\% & 82.2\% & 77.2\% \\
P1 & 600 & 62.8\% & 80.5\% & 85.5\% & 81.0\% & 76.3\% \\
P2 & 600 & 96.3\% & 98.0\% & 98.4\% & 98.1\% & 97.8\% \\
P3 & 600 & 98.2\% & 99.6\% & 99.5\% & 99.5\% & 99.2\% \\
\bottomrule
\end{tabular}
\label{tab:conditionresults}
\end{table}

The P1--P0 exact-set difference was 1.5 percentage points (95\% state-cluster bootstrap interval [--3.2, 6.3], Holm-adjusted $p=.6022$). In this study, explicitly asking models to reconstruct a prerequisite graph therefore did not provide evidence of better next-step selection. The F1 difference was slightly negative (--1.2 points), principally because recall decreased by 2.6 points.

In contrast, P2 exceeded P1 by 33.5 exact-accuracy points (95\% interval [22.7, 44.7], adjusted $p<.0001$) and 17.1 F1 points. P3 exceeded P2 by a further 1.8 exact points (95\% interval [0.5, 3.2]); its adjusted randomization $p=.0505$ was just above .05. The interval and adjusted test summarize the same small observed difference using different procedures; they do not establish equivalence between P2 and P3. All contrasts are interpreted descriptively, as specified above.

To answer RQ1, the largest observed gain was associated with providing the reference relation rather than asking models to reconstruct prerequisites. Explicit reconstruction did not show a clear aggregate advantage over descriptions alone, while both supplied-reference conditions achieved high accuracy. These results compare the implemented prompt conditions, including their differing output requirements; they do not isolate a single causal mechanism.

\subsubsection{RQ2: Variation Across Models, Domains, and Presentations}

\begin{figure}[ht]
\centering
\includegraphics[width=0.95\linewidth]{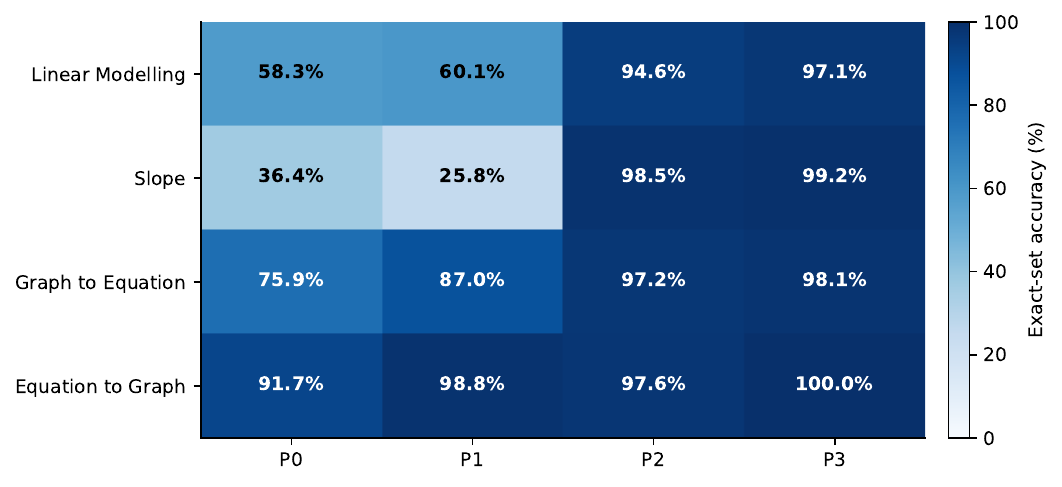}
\caption{Exact-set accuracy by domain and condition. Cell sizes differ because all unique states were included: 23, 11, 9, and 7 states per domain, respectively.}
\label{fig:domain}
\end{figure}

Accuracy differed substantially across domains (Figure~\ref{fig:domain}). Aggregated over conditions, models, and orders, exact-set accuracy was 97.0\% for Equation to Graph, 89.6\% for Graph to Equation, 77.5\% for Linear Modelling, and 65.0\% for Slope. The condition-by-domain heatmap (Figure~\ref{fig:domain}) shows that these differences were concentrated in P0 and P1. In P2, domain-specific exact accuracy ranged from 94.6\% to 98.5\%; in P3, it ranged from 97.1\% to 100\%. The observed differences across domains were therefore smaller when the reference structure was supplied.

Slope was the hardest ungrounded domain. Exact accuracy was 36.4\% in P0 and fell to 25.8\% in P1, before rising to 98.5\% in P2 and 99.2\% in P3. Linear Modelling was less extreme: exact accuracy was 58.3\% in P0, 60.1\% in P1, 94.6\% in P2, and 97.1\% in P3. The two representation-conversion domains were comparatively easy even without supplied structure.

Descriptive model exact accuracy ranged from 75.5\% to 86.0\% when averaged over all four conditions, compared with a 36.8-point range across conditions. P2 exceeded P1 for all six models. For selection within a particular application, condition-specific estimates are more informative than an average across prompts supplying different information. Table~\ref{tab:modelconditions} reports the model-by-condition estimates. GPT-5.6 Sol, Gemini 3.1 Pro Preview, and Gemini 3.7 Flash each reached 100\% exact accuracy in P2 on these cases, while their P0 values differed. These are observed scores for the included states, not guarantees for untested domains or future responses.

 \begin{table}[htbp]
	\centering
	\small
	\caption{Exact-set accuracy by model and prompt condition.
		Parentheses indicate ranks within each condition, pooled
		across domains. Tied scores receive the same rank, with
		subsequent ranks skipped. Each cell includes 100 responses
		across 50 domain-specific states and two presentations.}
	\label{tab:modelconditions}
	\begin{tabular}{lrrrr}
		\toprule
		Model & P0 & P1 & P2 & P3 \\
		\midrule
		Claude Fable 5
		& 76\% (1) & 70\% (1) & 98\% (4) & 100\% (1) \\
		Claude Sonnet 5
		& 60\% (4) & 61\% (4) & 85\% (6) & 96\% (5) \\
		GPT-5.6 Sol
		& 62\% (2) & 61\% (4) & 100\% (1) & 100\% (1) \\
		GPT-5.6 Terra
		& 57\% (5) & 63\% (2) & 95\% (5) & 93\% (6) \\
		Gemini 3.1 Pro Preview
		& 51\% (6) & 63\% (2) & 100\% (1) & 100\% (1) \\
		Gemini 3.7 Flash
		& 62\% (2) & 59\% (6) & 100\% (1) & 100\% (1) \\
		\bottomrule
	\end{tabular}
\end{table}

Both counterbalanced orders were available for all 1,200 model--condition--state pairs. Recommendation sets agreed exactly in 85.25\% of pairs, with a mean pairwise Jaccard similarity of 92.5\%. Thus, approximately one in seven pairs differed in at least one recommendation. Both orders were represented in every model--condition--state combination, ensuring balanced presentation coverage. However, the observed disagreement combines possible order effects with stochastic response variability; the design cannot separate them.

\begin{figure}[htbp]
	\centering
	\includegraphics[width=0.95\linewidth]{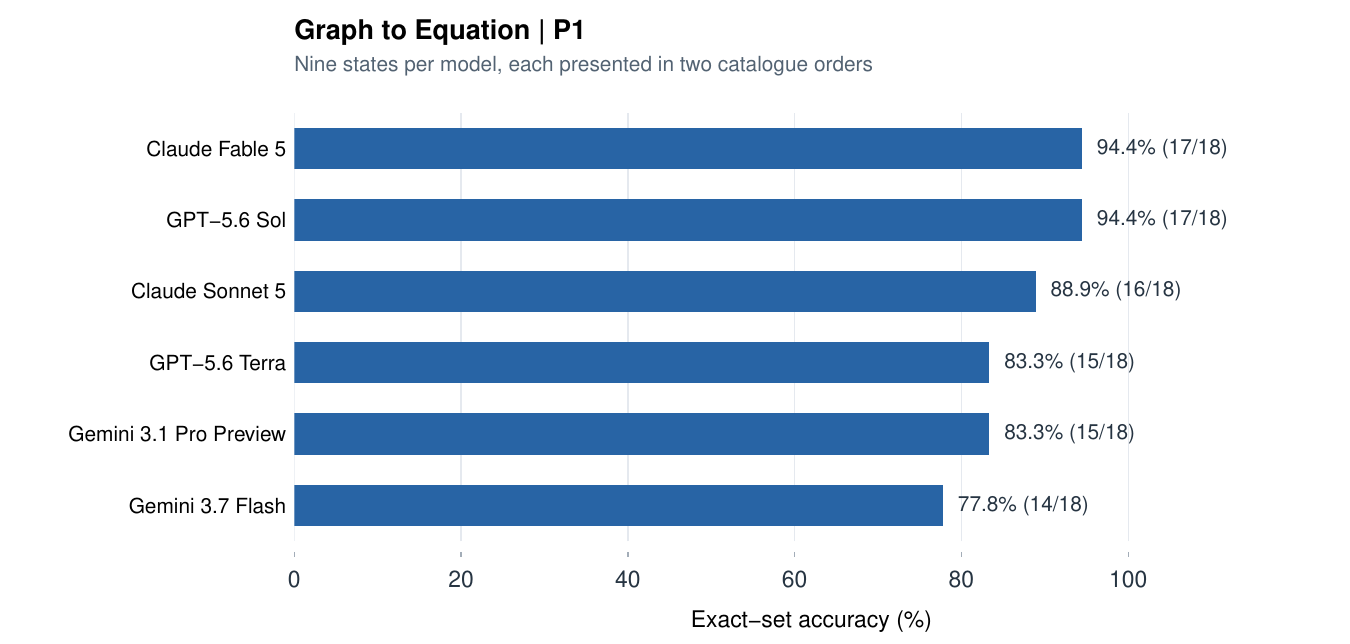}
	\caption{Model comparison for Graph to Equation under P1. Each model was evaluated on nine knowledge states presented in two catalogue orders. Labels show exact-set accuracy and the number of correct responses out of 18.}
	\label{fig:modelcomparisonp1}
\end{figure}

The comparison in Figure~\ref{fig:modelcomparisonp1} illustrates the benchmark's use for a specific selection decision. In Graph to Equation under P1, Claude Fable 5 and GPT-5.6 Sol achieved the highest observed exact-set accuracy, each recovering 17 of 18 reference sets (94.4\%). Claude Sonnet 5 recovered 16 sets, GPT-5.6 Terra and Gemini 3.1 Pro Preview each recovered 15, and Gemini 3.7 Flash recovered 14. Thus, models differed even within a domain in which aggregate performance was relatively high. These differences are descriptive: each model was evaluated on the same nine states under two presentations, and the difference between the highest and lowest scores corresponds to three responses.

The ordering of models changed across prompt conditions (Table~\ref{tab:modelconditions}). Claude Fable 5 ranked first in P0 and P1, but was overtaken in P2 by GPT-5.6 Sol and both Gemini models, which each achieved 100\% exact-set accuracy. Gemini 3.1 Pro Preview moved from the lowest observed accuracy in P0 to joint second place in P1 and joint first place in P2 and P3. Gemini 3.7 Flash likewise moved from the lowest score in P1 to joint first place in P2. In P3, four models shared the highest score. These results show both changes in relative performance and increasing ties near the accuracy ceiling when the reference structure was supplied.

Rankings also varied across domains within the same condition. Under P1, Claude Fable 5 had the highest exact-set accuracy in Linear Modelling (71.7\%), whereas Gemini 3.7 Flash led in Slope (36.4\%). In Graph to Equation, Claude Fable 5 and GPT-5.6 Sol shared the highest score (94.4\%); in Equation to Graph, five models achieved 100\%. Notably, Gemini 3.7 Flash ranked last when P1 results were pooled across domains, despite leading in Slope. Across all 16 domain--condition combinations, no model consistently achieved the highest score, even when shared first places were counted. A pooled ranking therefore does not identify the highest-performing model for every individual domain.

To answer RQ2, performance varied across both models and domains, and model comparisons depended on the prompt condition. Domain differences were concentrated in the conditions without a supplied reference structure. The two presentations usually produced the same set, but agreement was incomplete. The benchmark therefore provides distinct evidence about reference accuracy and agreement across presentations, rather than a single ranking that is independent of domain and prompt.

\section{Discussion}

\subsection{What the Benchmark Makes Measurable}

The methodological contribution is a reusable procedure for assessing whether a model--prompt combination recovers the admissible next-step set defined by an existing structured domain. KST supplies a reference that can be computed independently of the model and scored automatically. The prerequisite relation, task catalogue, state coverage, and prompt together specify the learning context of the benchmark score. This connects application-specific model selection to an explicit educational requirement: respecting the intended progression while recovering all admissible alternatives.

The sample study illustrates the value of this conditional interpretation. RQ1 showed much higher recommendation accuracy when the reference structure was supplied, while requesting prerequisite reconstruction did not show a clear aggregate advantage over descriptions alone. RQ2 showed that observed model performance depended on the prompt and domain, and that predictions did not always agree across presentations. These findings demonstrate the kinds of distinctions the benchmark can report. The approach complements evaluations of tutoring dialogue and pedagogical responses \parencite{macina2025mathtutorbench,kochmar2025findings} by isolating one formally specified selection task. It also addresses the contextual adaptivity concern raised by Borchers and Shou: recommendations can be evaluated against the progression intended by a particular application. Agreement is fidelity to that reference, rather than evidence that the reference is the only defensible curriculum \parencite{borchers2025adaptivity}.

\subsection{Using the Benchmark for Model and Prompt Selection}

An evaluator starts with the intended domain structure and recommendation task, specifies relevant state coverage, and compares candidate model--prompt combinations on the same cases. If the intended service supplies prerequisites, the corresponding condition is relevant to selection. If recommendations must be generated from descriptions alone, a descriptions-only condition addresses that requirement. P0--P3 are examples of such comparisons, not mandatory stages of the benchmark.

The model-by-condition results illustrate why an aggregate ranking across different prompts can be misleading for this purpose. Claude Fable 5 had the highest observed P0 accuracy, whereas three other models recovered every reference set in P2. In P3, four models reached 100\%, while GPT-5.6 Terra scored lower than in P2. A provider should therefore compare candidates under the information and output requirements of the intended application. Complete recovery on these finite cases does not guarantee error-free future responses.

Exact-set accuracy measures fulfillment of the complete-set task; precision and recall help distinguish additional inadmissible recommendations from omitted alternatives. Applications may attach different consequences to these errors, so selection also requires a justified acceptance criterion. Cost, latency, explanation quality, and other deployment requirements can then be assessed alongside structural performance. The present benchmark does not combine these considerations into an overall utility score.

P2 and P3 encode the same reference progression in different forms. Their small aggregate difference therefore concerns performance with different representations, rather than the educational superiority of one reference over another. A compact prerequisite relation avoids listing every admissible state in the prompt. For larger domains, the benchmark can also evaluate sampled states and compute their fringes directly from prerequisites; the present study does not establish the accuracy or cost of such larger-scale applications.

\subsection{A Benchmark Reference and Deterministic Recommendation}

If the prerequisite relation and the learner's knowledge state are already available, the complete admissible next-step set can be computed deterministically. An LLM is not needed for that set-selection operation. Indeed, this deterministic computation supplies the benchmark's reference answer. High scores in P2 or P3 consequently show how faithfully the tested models reproduce the reference under those conditions; they do not establish an advantage over directly computing it.

The benchmark remains useful when evaluating recommendations produced within an LLM-based service, including prompts that do not provide the relation. An application could also compute the admissible set deterministically and use an LLM to explain or contextualize the options. Whether that architecture improves usefulness or learning requires separate evaluation. The distinction between a computable selection rule and an evaluated language-model output keeps the contribution focused on measurement rather than on prescribing an LLM-based recommender.

\subsection{Limitations and Further Evaluation}

The benchmark evaluates agreement with a specified domain structure. Its interpretation therefore depends on the quality and educational meaning of that reference. Context-specific prerequisite relations allow the benchmark to reflect an intended instructional progression, but a high score does not establish that this progression is universally appropriate. Reference structures and their provenance must therefore be documented when the benchmark is applied. The present implementation uses quasi-ordinal knowledge spaces; extending it to alternative combinations of prerequisites would require richer KST representations.

The evaluation target is the complete set of admissible immediate next steps. This provides an explicit criterion for structural correctness, but does not determine which admissible option is pedagogically preferable, distinguish the educational severity of different errors, or establish the quality of accompanying explanations. Structural accuracy should therefore be considered one component of recommendation quality. Establishing usefulness and learning benefit requires additional evaluation involving learners and educational practice.

Benchmark scores also depend on the knowledge states included and their weighting. Exhaustive coverage describes performance across all admissible states of a given structure, whereas sampled or population-weighted evaluations address different questions. Larger domains may make exhaustive evaluation impractical, requiring a documented sampling procedure. Further evaluation should examine how structural characteristics, state coverage, and weighting affect the resulting estimates and model comparisons.

The empirical application is restricted to four small domains within linear functions, with at most seven components and 23 states in an individual domain. Although exhaustive coverage is feasible here, the results do not establish performance across larger structures, other subjects, languages, or curricular traditions. The presented study should therefore be understood as a worked example of how the benchmark can be applied to a specific use case. Broader applications are needed to assess its practical applicability and the interpretability of its scores across different settings.

Beyond extending empirical coverage, the benchmark provides a formal reference for systematically investigating the conditions under which LLMs generate correct next-step recommendations. Researchers can vary the information supplied in the prompt, the representation of prerequisite relations, and the presentation of the task while keeping the reference targets fixed. Comparisons across structures and knowledge states can additionally examine where performance deteriorates and whether the same conditions affect different models similarly. Repeated calls within each presentation would help distinguish stochastic variability from systematic presentation effects. Comparisons must also account for differences in output requirements, such as the additional graph requested in P1.

A planned qualitative analysis will examine the justifications accompanying the generated recommendations. The KST reference makes it possible to distinguish explanations associated with admissible recommendations from those supporting recommendations outside the outer fringe. The analysis can investigate which prerequisites models invoke, how they refer to the supplied knowledge state, and whether they introduce assumptions that are not part of the reference structure. Comparing these patterns across models, domains, and prompt conditions could reveal whether similar benchmark scores conceal different types of justification. Correct selection does not necessarily imply an appropriate explanation, and generated explanations should not be treated as direct evidence of a model's internal reasoning.

A complementary research direction concerns the relationship between reproducing a domain structure and achieving high recommendation accuracy. When a model returns an explicit prerequisite graph, its agreement with the reference relation can be examined alongside its benchmark score. Comparing the recommendation set with the fringe implied by the returned graph could further distinguish disagreement with the reference structure from inconsistency in applying the structure the model has produced. This relationship need not be one-to-one. Different prerequisite graphs can yield the same outer fringe for a particular knowledge state, while a single structural difference can affect recommendations across multiple states. The benchmark therefore enables analyses of which structural deviations are consequential for recommendation accuracy and where those consequences become visible. Such comparisons should distinguish differences in graph representation from differences in implied prerequisites, for example by examining transitive closures alongside direct relations, and account for the grouping of responses by model, domain, and knowledge state.

Together, these extensions would allow the benchmark to support both performance assessment and systematic investigation of the conditions associated with correct recommendations. Qualitative and structural analyses could then inform targeted revisions of prompts and domain representations, whose effects can be evaluated against the same explicit reference.

\section{Conclusion}

This paper presents a KST-based benchmarking method for evaluating LLM next-step recommendations against an existing domain structure. The method derives reference sets, evaluates specified model--prompt combinations on defined knowledge states, and reports exact and partial agreement. It provides a common scoring procedure that can be applied to different curricular references.

The sample study illustrates how this procedure reveals differences across prompt conditions, models, domains, and presentations. Supplying the reference structure was associated with substantially higher recommendation accuracy, but the results remain conditional on the tested cases and settings. The contribution is an explicit basis for assessing structural recommendation accuracy as one part of application-specific evaluation. It neither establishes learning benefit nor makes an LLM necessary where the admissible set can already be computed deterministically.

\section*{Statement on the Use of Artificial Intelligence}
\addcontentsline{toc}{section}{Statement on the Use of Artificial Intelligence}

Large language models are the objects of study and generated the analyzed recommendation corpus as described in the Method section. OpenAI Codex was additionally used to assist with data-collection and analysis code, consistency checks, and language drafting. The authors take full responsibility for the manuscript.

\section*{Declaration of Conflicting Interest}
\addcontentsline{toc}{section}{Declaration of Conflicting Interest}

The authors declared no potential conflicts of interest with respect to the research, authorship, or publication of this article.

\section*{Funding}
\addcontentsline{toc}{section}{Funding}

The work on this chapter was funded by the SNSF, grant number 100019\_219884. 

\phantomsection
\printbibliography[title={References}]

\end{document}